\documentclass[runningheads]{llncs}

\usepackage{todonotes}
\usepackage[T1]{fontenc}
\usepackage{xcolor}
\usepackage[colorlinks=true,linkcolor=black,anchorcolor=black,citecolor=black,filecolor=black,menucolor=black,runcolor=black,urlcolor=black]{hyperref}
\usepackage{graphicx}
\usepackage{booktabs}
\usepackage{graphicx}
\usepackage{caption}
\usepackage{subcaption}
\usepackage{layout}
\usepackage{wrapfig}
\usepackage{placeins}
\usepackage{orcidlink}
\usepackage{fontawesome5}
\usepackage{hyphenat}
\usepackage[normalem]{ulem}
\usepackage{soul}

\begin{document}

\title{Faithful Faithfulness Evaluations: Challenges \& Pitfalls Learned from a Breast MRI Case Study}
\titlerunning{Faithful Faithfulness Evaluations}

\author
{
Peachapong Poolpol\inst{1,2}\orcidlink{0000-0002-8840-8785} \and
Henrik H.J. Detjen\inst{1}\orcidlink{0000-0002-7683-1797} \and
Eike Petersen\inst{1}\orcidlink{0000-0003-0097-3868}
}

\authorrunning{P. Poolpol et al.}

\institute{
Fraunhofer Institute for Digital Medicine MEVIS, Bremen, Germany
\and Deggendorf Institute of Technology, 
Deggendorf, Germany
\\\email{\{eike.petersen,henrik.detjen\}@mevis.fraunhofer.de, \\peachapong.p@gmail.com}}

\maketitle
    
\begin{abstract}
    Saliency maps are widely used to explain deep learning predictions in medical imaging, yet visually plausible explanations do not necessarily reflect a model's true decision process and may therefore mislead clinicians. We investigate this problem using a Vision Transformer-based breast MRI classifier trained on the ODELIA Breast MRI Challenge dataset and evaluate multiple saliency methods, including Last-layer Attention, Attention Rollout, Grad-SAM, Gradient Attention Rollout, GMAR, Grad-CAM, and HiResCAM. Our study highlights two often-overlooked challenges in perturbation-based faithfulness evaluation. First, method rankings depend strongly on the perturbation strategy, varying across intensity-based perturbations and transformer-based attention masking. Second, benchmarking saliency methods requires distinguishing between class-specific and class-agnostic explanations. To enable fair comparisons, we introduce non-class-specific variants of gradient-based methods and evaluate both settings separately. Across protocols, Grad-CAM and Gradient Attention Rollout consistently emerged as the strongest class-specific methods, although their relative ranking depended on the evaluation design. These findings expose important limitations of current saliency-based explainability approaches and highlight the need for more robust and standardized evaluation frameworks for trustworthy clinical AI systems.
\end{abstract}

\begin{figure}[h]
\centering
\vspace{-0.5cm}
\includegraphics[width=\textwidth]{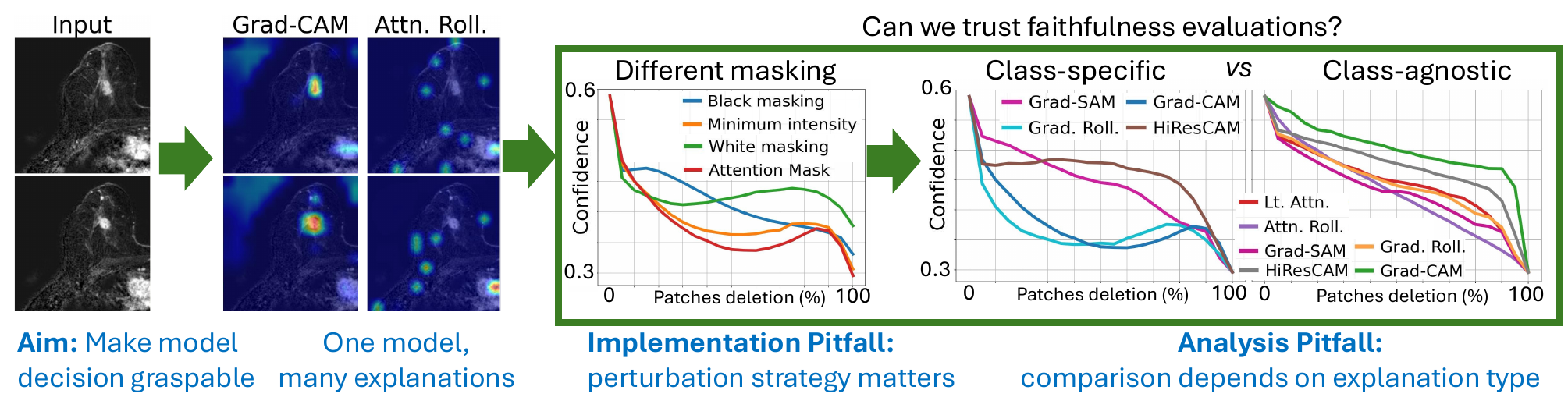}
\caption{Key pitfalls in saliency faithfulness evaluation –- we discuss recommendations derived from our use case for more reliable faithfulness assessment.}
\vspace{-1.0cm}
\label{fig:graphical-abstract}
\end{figure}


\section{Introduction}

Post-hoc explanation methods are frequently used to interpret deep learning models in medical imaging. In image classification, these methods commonly produce saliency maps, highlighting regions assumed to be relevant for a prediction and offering an intuitive visual explanation.  
However, visual plausibility alone is not sufficient for judging explanation quality: 
a visually plausible heatmap may suggest that the model focuses on a lesion, yet quantitative testing may reveal that perturbing the highlighted region has little effect on the prediction~\cite{adebayo2018sanity,hooker2019benchmark,Zhang2024a}. Conversely, an explanation may be quantitatively faithful but clinically uninformative if the model relies on confounders. Therefore, faithfulness evaluation should be treated as a separate methodological assessment.

Perturbation-based evaluation has become a common approach for assessing explanation faithfulness~\cite{Zhang2024a,Barekatain2025,Kashefi2023}. By modifying image regions according to a saliency ranking and measuring the resulting prediction changes, these methods estimate how strongly highlighted regions influence model behavior. However, evaluation outcomes can depend on methodological choices such as the perturbation strategy and evaluation protocol, see figure~\ref{fig:graphical-abstract}.
Here, we study the faithfulness of saliency maps in breast MRI classification. 
Our contributions include
\begin{enumerate}
    \item evaluating the effect of different perturbation strategies on faithfulness evaluations, demonstrating that explanation ranking may depend on this choice,
    \item a generic scheme for deriving class-agnostic saliency maps from class-specific methods, enabling between-family comparisons, and
    \item the first comprehensive evaluation of the faithfulness of seven saliency map methods in a ViT-based volumetric image classification setting. 
\end{enumerate}
\section{Background \& Related Work}
\subsection{Saliency-Based Explanation Methods}

Gradient-based and attention-based methods are the most commonly used post-hoc explanation approaches for image classifiers. Grad-CAM uses gradients of a target class with respect to internal feature activations to generate class-discriminative localization maps \cite{selvaraju2017gradcam} 
and has been adapted to transformer-based vision models by treating patch tokens as spatial feature representations~\cite{Barekatain2025,Kashefi2023}. HiResCAM~\cite{draelos2021usehirescaminsteadgradcam} extends Grad-CAM by replacing gradient averaging with element-wise multiplication between activations and gradients, aiming to improve explanation faithfulness while retaining class-discriminative localization.

For Vision Transformers (ViTs)~\cite{dosovitskiy2021an}, attention weights are often used as explanations because they directly represent token interactions. Last-layer attention visualizes the attention from the classification token to image patches. However, raw attention provides only a partial view of information flow and is not necessarily equivalent to feature importance \cite{jain2019attention}. Attention rollout addresses this limitation by recursively combining attention matrices across layers to approximate how information propagates through the transformer \cite{abnar2020quantifying}, yet this approach remains empirically motivated and not theoretically justified. 

More recent transformer-specific explanation methods combine attention information with gradient information to obtain class-specific attributions~\cite{chefer2021transformer}. Grad-SAM weights self-attention maps using class gradients to generate gradient-guided attention explanations~\cite{Barkan2021Grad-SAM}. Gradient Attention Rollout extends attention rollout by incorporating gradient information into the attention aggregation process, producing class-specific explanations while preserving attention flow across layers~\cite{Barekatain2025}. Similarly, GMAR combines gradient weighting with multi-head attention rollout and has been proposed as an alternative transformer explanation strategy~\cite{Jo2025GMAR}. 

\subsection{Faithfulness of Saliency Maps}

A central concern in explainable AI is whether explanations are \emph{faithful}: an explanation should identify all image regions that substantially influence the model's output, and no other regions.
Prior work has shown that visual inspection is insufficient for this purpose. 
Adebayo et al.~\cite{adebayo2018sanity} introduced sanity checks demonstrating that saliency methods can be insensitive to model parameters or labels. 
Hooker et al.~\cite{hooker2019benchmark} proposed a benchmark based on removing high-saliency features and retraining models, showing that several attribution methods perform close to random rankings.
However, subsequent work demonstrated that fixed-value perturbations can confound evaluation results, motivating in-distribution perturbation strategies such as ROAD~\cite{Rong2022ROAD}.
Zhang et al.~\cite{Zhang2024a} and Dombrowski et al.~\cite{Dombrowski2019} showed that saliency maps are sensitive to imperceptible adversarial changes in input images, raising concerns about their reliability.

Perturbation-based faithfulness evaluations assess faithfulness by modifying features according to their saliency values. Deletion and insertion curves were popularized by Petsiuk et al.~\cite{petsiuk2018rise} as quantitative measures. However, perturbation methods are sensitive to design choices including perturbation granularity, replacement strategy, and whether confidence is measured for the predicted or the ground-truth class. 
For ViTs specifically, masking should be conducted using attention masking instead of fixed-value replacement or noising~\cite{Kashefi2023,Mehri2024}, but this is not yet commonly done within the medical imaging community~\cite{Barekatain2025}.

\subsection{Explainability in Medical Imaging}

Saliency maps are often interpreted as evidence localization~\cite{Mueller-Franzes2025MST}, yet clinical plausibility and model faithfulness are distinct. A heatmap may overlap with a lesion because of image structure, preprocessing, or inductive bias, without necessarily reflecting the evidence used by the model.
Prior work on transformer explanations in medical imaging has reported inconsistent findings across tasks and architectures, suggesting that explanation quality may depend on pathology, image modality, and evaluation protocol.
Most recently, Barekatain and Glocker~\cite{Barekatain2025} evaluated the faithfulness of ViT explanation strategies in a medical imaging setting, but their analysis was limited to two explanation methods and 2D images; a comprehensive evaluation of ViT saliency faithfulness in a volumetric setting is currently lacking.
\section{Methods}
\subsection{Data and Prediction Setting}

Experiments were conducted on the public training and validation portion of the ODELIA Breast MRI Challenge 2025 dataset~\cite{MuellerFranzes2025}. The dataset contains 1,022 unilateral breast MRI samples labeled as No-lesion, Benign, or Malignant. Data were split at patient level into 792 training, 114 validation, and 116 test samples. Classification was performed using a Medical Slice Transformer (MST)~\cite{Mueller-Franzes2025MST}, which processes 3D MRI volumes as sequences of 2D slices. Slice features were extracted using a pretrained DINOv3-ViT~\cite{Simoni2025DINOv3} backbone and aggregated by a slice-fusion transformer to produce three-class predictions. The model was used only for explanation evaluation; model optimization was not a focus of our study. 

\subsection{Explanation Methods}

For each test sample, saliency maps were generated using Last-layer Attention~\cite{jain2019attention}, Attention Rollout~\cite{abnar2020quantifying}, Grad-SAM~\cite{Barkan2021Grad-SAM}, Gradient Attention Rollout~\cite{Barekatain2025}, GMAR~\cite{Jo2025GMAR} (L1 and L2 gradient norm), Grad-CAM~\cite{selvaraju2017gradcam}, and HiResCAM~\cite{draelos2021usehirescaminsteadgradcam}. Since Last-layer Attention and Attention Rollout operate at the slice level in the MST, saliency maps were weighted by slice-fusion transformer attention scores. Gradient-based methods generate explanations directly from the final prediction and therefore require no additional slice weighting. 
To enable comparison with class-agnostic explanations, we additionally constructed non-class-specific variants of class-specific explanations by averaging saliency maps across output classes. For these non-class-specific variants only, absolute gradients were used instead of ReLU to aggregate positive and negative contributions across classes. 

\subsection{Perturbation-Based Faithfulness Evaluation}
Saliency maps were converted to patch-level importance scores via interpolation.
In deletion, starting from the original image, perturbations are applied in descending order of patch importance; more faithful explanations should cause a faster confidence decrease and lower area under the confidence curve (AUCC). In insertion, starting from an empty image, the most salient patches are progressively restored;  more faithful explanations should produce faster confidence recovery and higher AUCC values.

\subsection{Perturbation Strategies}

To assess the sensitivity of faithfulness evaluations to the perturbation strategy, we evaluated black masking, white masking, minimum-intensity replacement, and attention masking. As inputs were z-score normalized, fixed-intensity masking used replacement values $z=-10$ (`black') and $z=10$ (`white'), outside the observed test-set intensity range.
Minimum-intensity replacement used the minimum intensity of each image. Attention masking~\cite{Vaswani2017} suppresses selected patches directly within the transformer's attention computation, thus fully removing a patch from model computations instead of replacing it with a fixed value.

\subsection{Aggregation and Analysis}

For each explanation method and perturbation setting, confidence was recorded in 5\% perturbation steps. Perturbation curves were summarized using AUCC. For each class, AUCC values were computed on a per-sample basis, from which the mean and 95\% confidence interval (mean $\pm$ 1.96 SEM) were estimated. Macro-averaged results were then obtained by averaging class-wise means and corresponding confidence interval bounds.
In addition to full AUCC, we report AUCC over the first 30\% of perturbation steps to emphasize early confidence changes, corresponding to the highest-saliency image regions. For visualization, confidence curves were obtained by averaging confidence values across test samples at each perturbation step.
Our code is available at
\url{https://github.com/friendorus/Faithful-Faithfulness-Evaluations}.
\section{Results}

\subsection{Baseline Prediction Performance}

The model achieved moderate test-set classification performance (accuracy $0.57$, macro $F_1$ $0.52$, weighted $F_1$ $0.57$, macro AUROC $0.73$, micro AUROC $0.78$). Performance varied across classes, with the highest one-vs-rest AUROC observed for the -- arguably most important -- malignant cases ($0.81$)(precision $0.77$, recall $0.53$, $F_1$ $0.63$), followed by no-lesion ($0.69$) and benign ($0.69$). Since the focus of this work is the evaluation of explanation faithfulness, these results are provided primarily as context for interpreting the subsequent saliency analyses.

\subsection{Qualitative Comparison of Explanation Methods}

\begin{figure*}[t]

\begin{subfigure}[b]{\textwidth}
(a)
    \includegraphics[height=1.4cm]{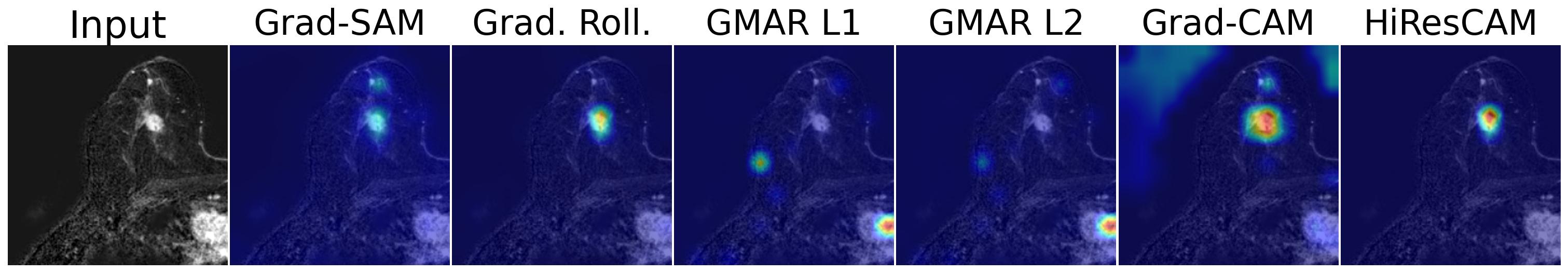}
    \label{fig:class_malignant}
\end{subfigure}

\begin{subfigure}[b]{\textwidth}
(b)
    \includegraphics[height=1.4cm]{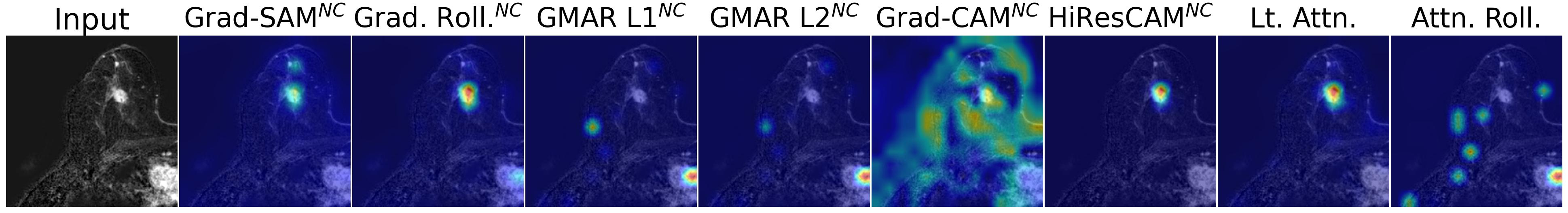}
    \label{fig:nonclass_malignant}
\end{subfigure}

\begin{subfigure}[b]{\textwidth}
(c)
    \includegraphics[height=1.4cm]{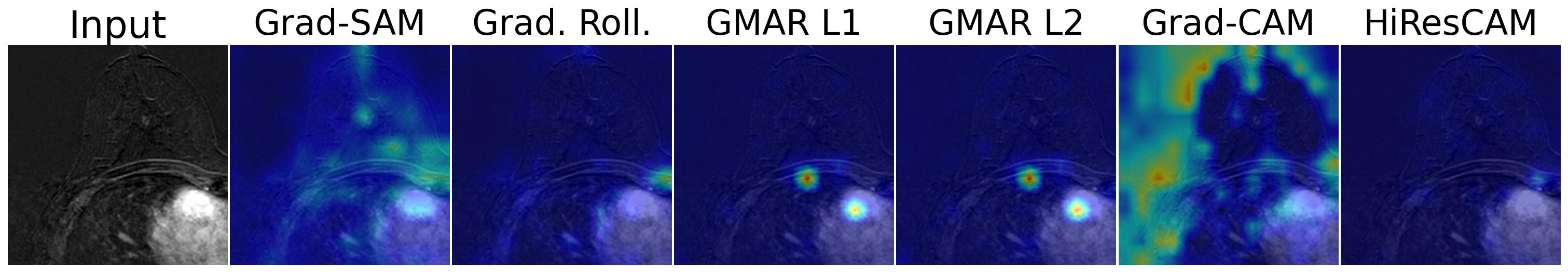}
    \label{fig:class_nolesion}
\end{subfigure}

\begin{subfigure}[b]{\textwidth}
(d)
    \includegraphics[height=1.4cm]{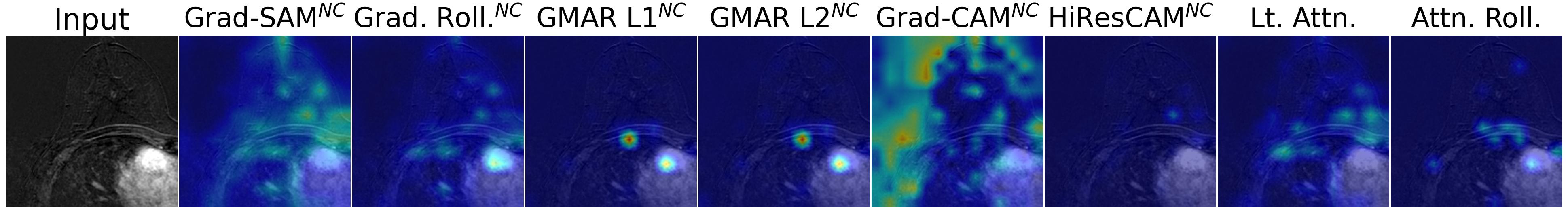}
    \label{fig:nonclass_nolesion}
\end{subfigure}
\caption{Saliency maps of a Malignant case ((a) class-specific, (b) non-class-specific) and a No-lesion case ((c) class-specific, (d) non-class-specific).}
\label{fig:qualitative_examples}

\vspace{1mm}
\raggedright
\scriptsize
\textit{Note.} Grad. Roll. = Gradient Attention Rollout, Lt. Attn. = Last-layer Attention, \\Attn. Roll. = Attention Rollout, NC = non-class-specific methods
\end{figure*}

Figure~\ref{fig:qualitative_examples} shows saliency maps for malignant and no-lesion cases. For the malignant case, all class-specific methods highlight the lesion region, although Grad-CAM produces more diffuse saliency than the other methods. Most non-class-specific methods also localize the lesion, whereas Attention Rollout and Grad-CAM generate more dispersed saliency. Saliency maps for the no-lesion case were more heterogeneous across methods, reflecting the absence of a localized target region. These qualitative observations motivate the quantitative faithfulness evaluation presented in the following sections.

\subsection{Effect of Replacement Strategy}

\begin{wrapfigure}{r}{0.43\textwidth}
    \centering
    \vspace{-0.9cm}
    \includegraphics[width=0.43\textwidth]{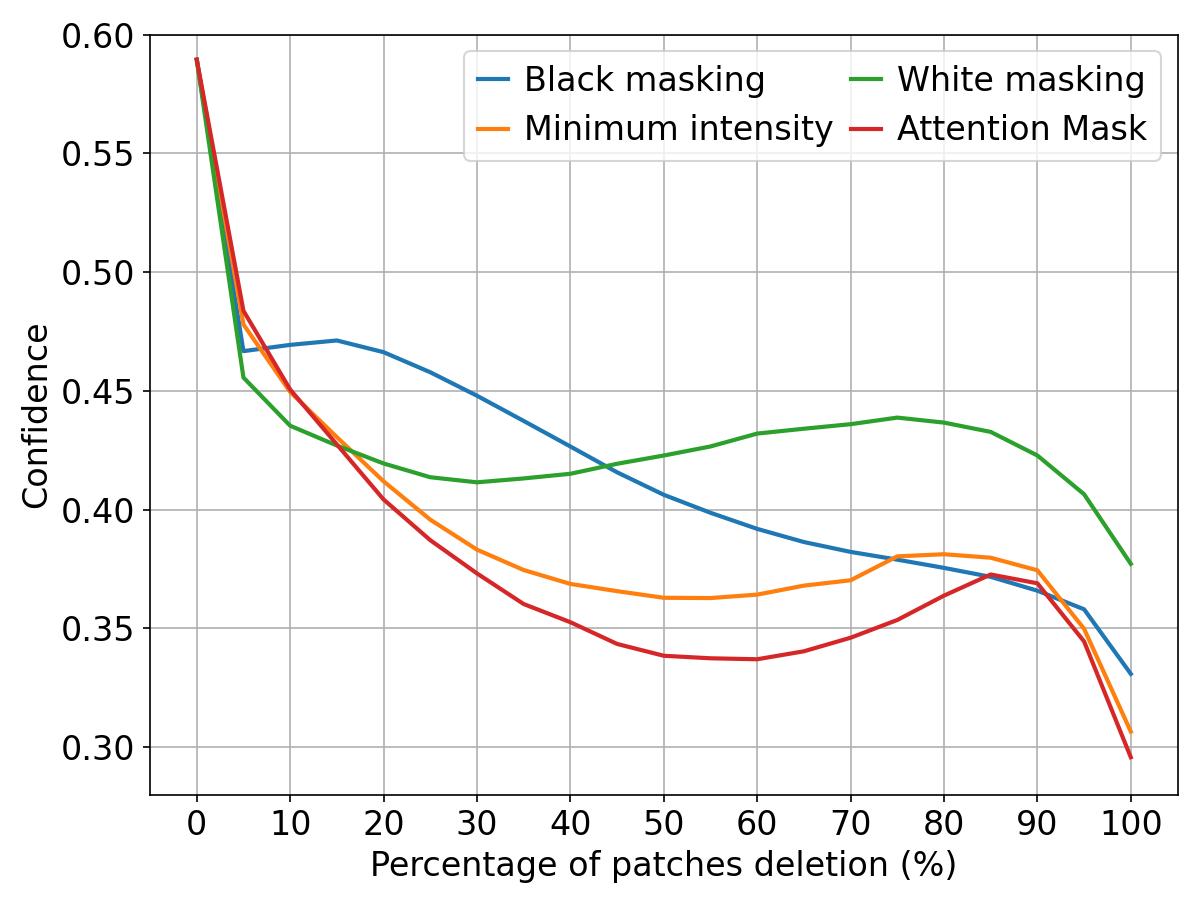}
    \vspace{-0.5cm}
    \caption{Deletion confidence curves for Grad-CAM under different replacement strategies.}
    \label{fig:replacement_strategy_gradcam}
    \vspace{-0.5cm}
\end{wrapfigure}

Replacement strategy substantially affected perturbation-based faithfulness (Table~\ref{tab:replacement_strategy}). While Gradient Attention Rollout and Grad-CAM consistently achieved the strongest deletion performance, their relative ranking depended on the replacement strategy. Figure~\ref{fig:replacement_strategy_gradcam} further illustrates that different replacement strategies produce distinct perturbation curves even for the same explanation method. Because it constitutes the most principled removal strategy, attention masking was selected for subsequent experiments.

\begin{table*}[t]
\centering
\caption{Macro-averaged deletion AUCC ($\downarrow$) [95\% CI] for class-specific explanation methods evaluated using different replacement strategies.}
\label{tab:replacement_strategy}
\small
\begin{tabular}{lcccc}
\toprule
\textbf{Method}
& \textbf{Black}
& \textbf{Min. Intensity}
& \textbf{White}
& \textbf{Attn. Mask} \\
\midrule
Random
& 0.355 [.348,.363]
& 0.476 [.438,.514]
& 0.409 [.383,.435]
& 0.512 [.479,.546] \\

Grad-SAM
& 0.352 [.346,.358]
& 0.377 [.362,.391]
& 0.373 [.358,.389]
& 0.412 [.394,.430] \\

Grad. Roll.
& \textbf{0.346} [.340,.353]
& \textit{0.327} [.311,.342]
& \textit{0.342} [.327,.357]
& \textbf{0.336} [.318,.355] \\

GMAR-L1
& 0.355 [.347,.362]
& 0.376 [.361,.392]
& 0.369 [.353,.385]
& 0.412 [.394,.430] \\

GMAR-L2
& 0.355 [.347,.362]
& 0.376 [.361,.391]
& 0.369 [.353,.385]
& 0.412 [.393,.430] \\

Grad-CAM
& \textit{0.347} [.338,.356]
& \textbf{0.323} [.304,.342]
& \textbf{0.332} [.314,.350]
& \textbf{0.336} [.314,.358] \\

HiResCAM
& 0.358 [.351,.366]
& 0.388 [.366,.410]
& 0.383 [.362,.403]
& 0.417 [.390,.444] \\

\bottomrule
\end{tabular}

\vspace{1mm}
\raggedright
\scriptsize
\textit{Note.} Grad. Roll. = Gradient Attention Rollout
\end{table*}

\subsection{Deletion and Insertion}

\begin{figure*}[t]
\centering

\begin{subfigure}[b]{0.495\textwidth}
    \includegraphics[width=\textwidth]{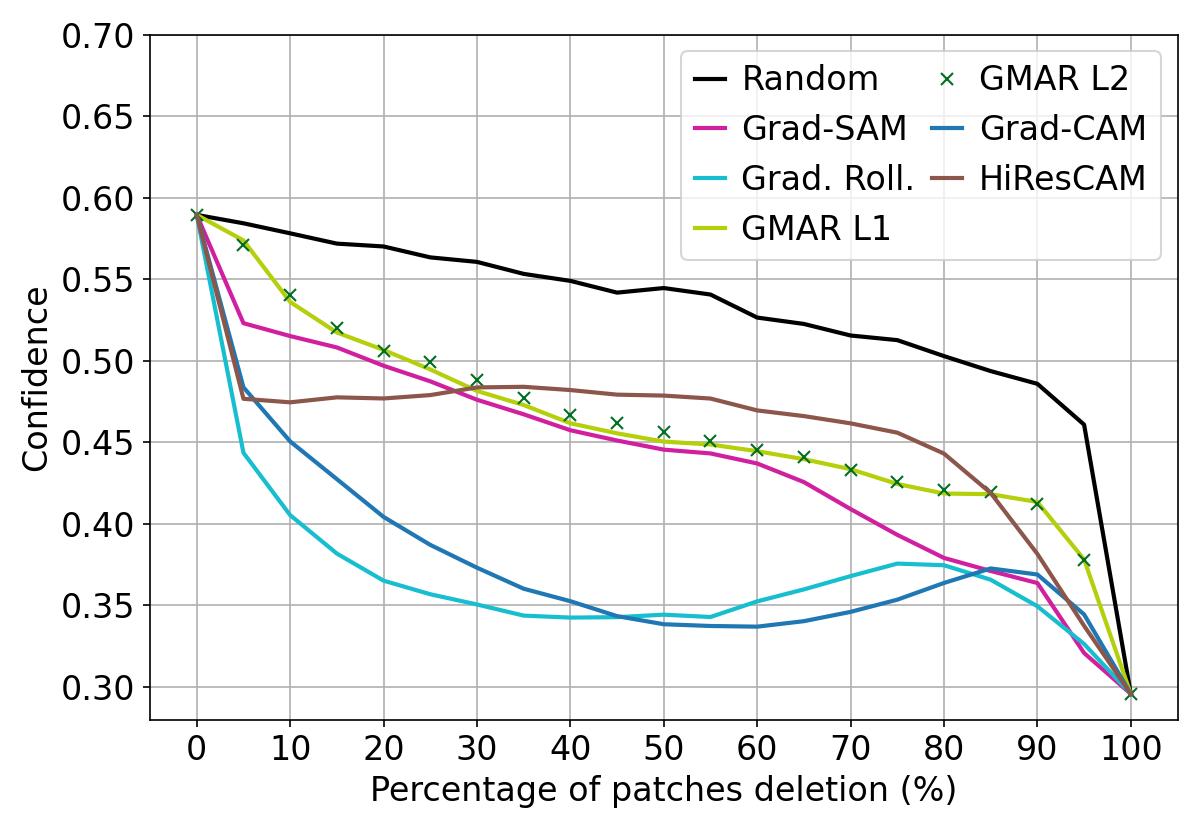}
    \caption{Deletion, class specific methods}
    \label{fig:del_class}
\end{subfigure}
\hfill
\begin{subfigure}[b]{0.495\textwidth}
    \includegraphics[width=\textwidth]{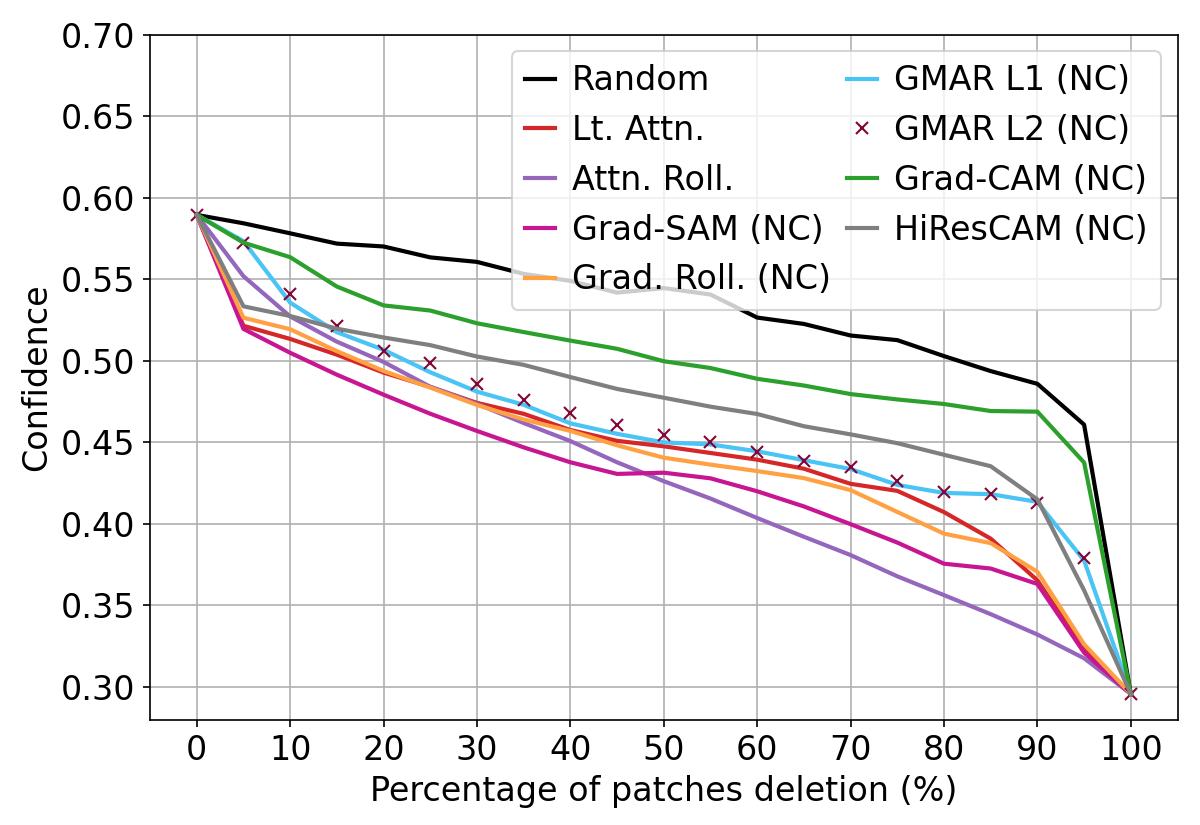}
    \caption{Deletion, non-class specific methods}
    \label{fig:del_nonclass}
\end{subfigure}

\vspace{0.5em}

\begin{subfigure}[b]{0.495\textwidth}
    \includegraphics[width=\textwidth]{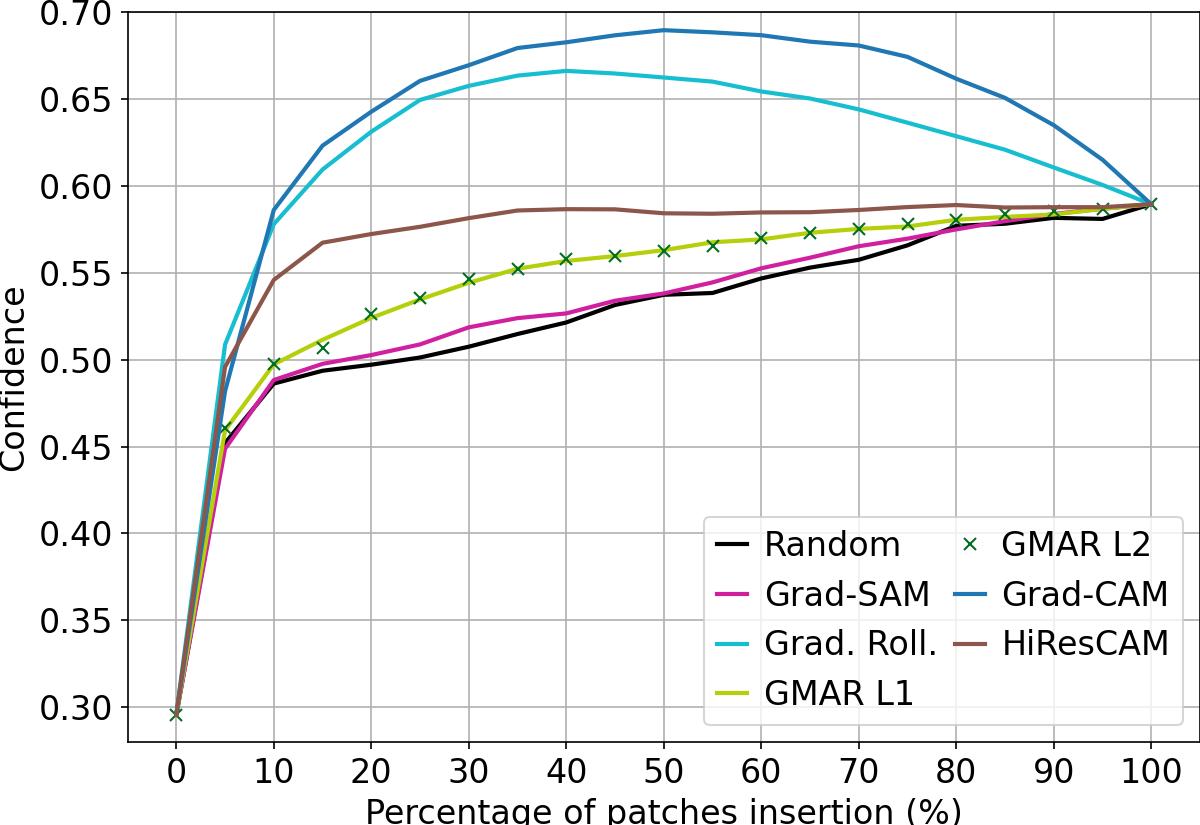}
    \caption{Insertion, class specific methods}
    \label{fig:ins_class}
\end{subfigure}
\hfill
\begin{subfigure}[b]{0.495\textwidth}
    \includegraphics[width=\textwidth]{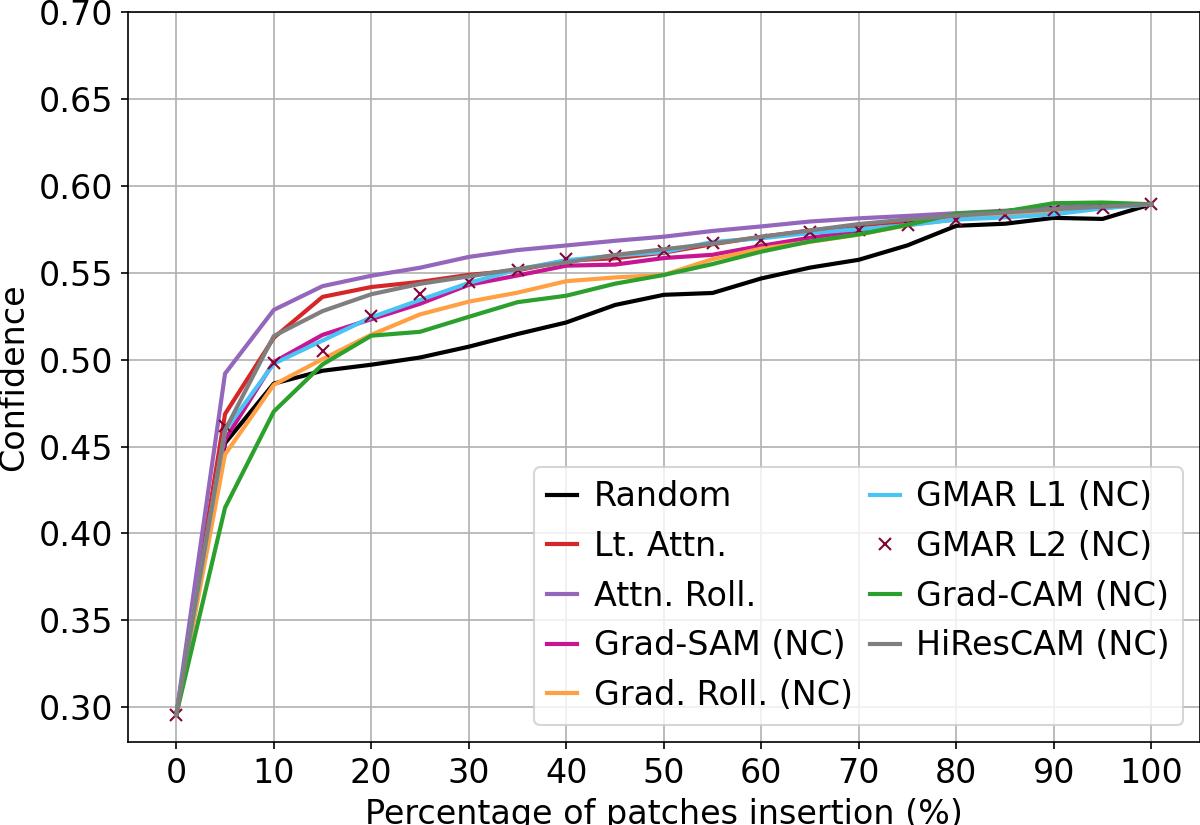}
    \caption{Insertion, non-class specific methods}
    \label{fig:ins_nonclass}
\end{subfigure}

\caption{
Confidence curves obtained from deletion and insertion evaluations using the attention-mask replacement strategy.
}
\label{fig:perturbation_curves}
\vspace{1mm}
\raggedright
\scriptsize
\textit{Note.} Grad. Roll. = Gradient Attention Rollout,  Lt. Attn. = Last-layer Attention, 

Attn. Roll. = Attention Rollout, NC = non-class specific methods
\end{figure*}

Table~\ref{tab:macro_aucc} summarizes AUCC scores obtained using attention masking; figure~\ref{fig:perturbation_curves} shows the corresponding confidence curves.
For class-specific explanations, Grad-CAM and Gradient Attention Rollout achieved the strongest performance, with Grad-CAM obtaining the lowest deletion AUCC and highest insertion AUCC, and Gradient Attention Rollout achieving the best Deletion$_{30}$ score.
Class-specific Grad-CAM and Gradient Attention Rollout outperformed their non-class-specific counterparts, whereas GMAR showed minimal differences. For non-class-specific explanations, performance differences were generally smaller, with Grad-SAM achieving the strongest deletion performance, HiResCAM the highest insertion scores, and Gradient Attention Rollout the best Deletion$_{30}$ score.

\begin{table*}[t]
\centering
\caption{Macro-averaged AUCC [95\% CI] for class-specific and non-class-specific explanation methods.}
\label{tab:macro_aucc}

(a) Class-specific methods
\small
\begin{tabular}{p{2.1cm} p{2.4cm} p{2.4cm} p{2.4cm} p{2.4cm}}
\toprule
\textbf{Method} &
\textbf{Deletion ($\downarrow$}) &
\textbf{Deletion$_{30}$ ($\downarrow$}) &
\textbf{Insertion ($\uparrow$}) &
\textbf{Insertion$_{30}$ ($\uparrow$}) \\
\midrule

Random &
0.512 [.479,.546] &
0.173 [.161,.185] &
0.505 [.471,.539] &
0.126 [.117,.135] \\

Grad-SAM &
0.412 [.394,.430] &
0.142 [.133,.150] &
0.525 [.486,.564] &
0.136 [.124,.147] \\

Grad. Roll. &
\textbf{0.336} [.318,.355]&
\textbf{0.110} [.103,.118] &
\textit{0.621} [.591,.652] &
\textit{0.167} [.158,.177] \\

GMAR-L1 &
0.445 [.413,.476] &
0.151 [.140,.162] &
0.531 [.493,.569] &
0.135 [.125,.145] \\

GMAR-L2 &
0.447 [.415,.479] &
0.152 [.141,.162] &
0.530 [.492,.568] &
0.135 [.125,.145] \\

Grad-CAM &
\textbf{0.336} [.314,.358] &
\textit{0.117} [.108,.127] &
\textbf{0.637} [.605,.669] &
\textbf{0.172} [.160,.183] \\

HiResCAM &
0.417 [.390,.444] &
0.133 [.123,.143] &
0.583 [.546,.619] &
0.165 [.152,.177] \\

\bottomrule
\end{tabular}

\vspace{2mm}

(b) Non-class-specific methods
\small
\begin{tabular}{p{2.1cm} p{2.4cm} p{2.4cm} p{2.4cm} p{2.4cm}}
\toprule
\textbf{Method} &
\textbf{Deletion ($\downarrow$}) &
\textbf{Deletion$_{30}$ ($\downarrow$}) &
\textbf{Insertion ($\uparrow$}) &
\textbf{Insertion$_{30}$ ($\uparrow$}) \\
\midrule

Random &
0.512 [.479,.546] &
0.173 [.161,.185] &
0.505 [.471,.539] &
0.126 [.117,.135] \\

Lt. Attn. &
\textit{0.407} [.390,.425] &
\textit{0.138} [.130,.146] &
0.562 [.524,.601] &
\textit{0.155} [.143,.168] \\

Attn. Roll. &
0.413 [.392,.433] &
0.144 [.134,.153] &
\textit{0.565} [.527,.602] &
\textit{0.155} [.144,.166] \\

Grad-SAM$^{NC}$ &
\textbf{0.403} [.386,.420] &
\textit{0.138} [.130,.145] &
0.539 [.500,.577] &
0.140 [.129,.152] \\

Grad. Roll.$^{NC}$ &
0.409 [.392,.425] &
\textbf{0.137} [.130,.144] &
0.537 [.500,.573] &
0.137 [.127,.147] \\

GMAR-L1$^{NC}$ &
0.444 [.413,.475] &
0.151 [.140,.161] &
0.531 [.493,.569] &
0.135 [.125,.145] \\

GMAR-L2$^{NC}$ &
0.447 [.415,.479] &
0.152 [.141,.162] &
0.530 [.492,.568] &
0.135 [.125,.145] \\

Grad-CAM$^{NC}$ &
0.475 [.445,.504] &
0.160 [.149,.172] &
0.528 [.488,.569] &
0.136 [.124,.147] \\

HiResCAM$^{NC}$ &
0.421 [.400,.443] &
0.141 [.133,.150] &
\textbf{0.568} [.529,.606] &
\textbf{0.158} [.145,.170] \\

\bottomrule
\end{tabular}

\vspace{1mm}
\raggedright
\scriptsize
\textit{Note.} Lt. Attn. = Last-layer Attention, Attn. Roll. = Attention Rollout, 

Grad. Roll. = Gradient Attention Rollout, NC = non-class-specific methods
\end{table*}

\section{Discussion}

Our study demonstrates that saliency map faithfulness evaluation is protocol-dependent. Method rankings varied across replacement strategies, deletion and insertion tests, and full versus early perturbation analyses. For example, Grad-CAM achieved the strongest overall deletion performance, whereas Gradient Attention Rollout showed stronger performance when considering only the highest-saliency regions. Consistent with Barekatain and Glocker~\cite{Barekatain2025}, Grad-CAM and Gradient Attention Rollout emerged as the strongest class-specific methods. 
The replacement strategy substantially affected faithfulness ranking. Under fixed-intensity replacement, explanation methods perform similarly to random masking, likely due to strong out-of-distribution artifacts. 
In contrast, attention masking produced greater separation between methods while preserving the original image intensities, providing a more reliable faithfulness evaluation.

Qualitative analysis further showed that visual appearance and quantitative evaluation are not equivalent. Most class-specific methods localized the lesion region in malignant cases, yet quantitative results differed substantially. Conversely, Grad-CAM and Gradient Attention Rollout achieved similar quantitative performance despite producing noticeably different saliency maps in no-lesion cases. Similar observations were reported by Wollek et al.~\cite{Wollek2023Attention-basedClassification}, who found that attention-based saliency maps achieved favorable quantitative performance and were considered useful by radiologists. 
While perturbation tests assess the relationship between explanations and model predictions, clinicians are primarily interested in whether explanations correspond to plausible disease-related features -- risking conflating clinical plausibility with explanation faithfulness.

Interpretation of attention-based methods also requires consideration of their inherently non-class-specific nature. Unlike gradient-based methods, Last-layer Attention and Attention Rollout are not tied to a specific output class. To enable a fair comparison, we introduced non-class-specific variants of the gradient-based methods and evaluated both settings separately. Under this comparison, performance differences became substantially smaller, suggesting that part of the advantage of class-specific methods arises from access to class-discriminative information rather than the explanation mechanism itself. The proposed non-class-specific variants are intended as a simple means to compare class-specific and class-agnostic methods; we do not claim optimality of this approach. Still, the derived variants achieved highly competitive performance relative to attention-based methods, suggesting that aggregation retained meaningful information.

Our study is limited to a single breast MRI classification task and one transformer architecture with moderate classification performance on a highly challenging diagnostic task. It remains unclear whether the ranking of saliency methods generalizes to stronger models. However, explanation methods should faithfully reflect the model's decision-making process regardless of predictive performance, as faithfulness is defined with respect to the learned model rather than the underlying task. In addition, although the overall performance was moderate, the model showed the strongest discriminative performance for the malignant class.
Future work should extend this framework across
imaging tasks, architectures, and explanation methods while investigating how perturbation-based evaluation relates to clinician trust and clinical utility.
\section{Conclusion}

We present a perturbation-based framework for evaluating saliency maps in transformer-based breast MRI classification. Our results show that explanation method rankings depend strongly on the evaluation protocol, including the choice of replacement strategy, perturbation metric, and class specificity. 
No single method was consistently superior across all settings. 
Our findings emphasize that saliency maps should be quantitatively evaluated before clinical interpretation and that explanation quality cannot be assessed by judging clinical plausibility.
\begin{credits}
\subsubsection{\ackname} This work has received funding from the European Union’s Horizon Europe research and innovation programme under grant agreement No 101057091. Views and opinions expressed are however those of the author(s) only and do not necessarily reflect those of the European Union or the European Health and Digital Executive Agency (HADEA). Neither the European Union nor the granting authority can be held responsible for them.

\subsubsection{\discintname}
The authors declare no competing interests.
\end{credits}
\FloatBarrier
\bibliographystyle{splncs04}
\bibliography{bibliography/references}

\end{document}